\documentclass[10pt,twocolumn,letterpaper]{article}

\usepackage{cvpr}
\usepackage{times}
\usepackage{epsfig}
\usepackage{graphicx}
\usepackage{amsmath}
\usepackage{amssymb}
\usepackage{multirow}
\usepackage{verbatim}
\usepackage{appendix}
\usepackage{subcaption}

\usepackage[pagebackref=true,breaklinks=true,letterpaper=true,colorlinks,bookmarks=false]{hyperref}

\cvprfinalcopy 

\def\cvprPaperID{386} 

\newcommand\blfootnote[1]{%
  \begingroup
  \renewcommand\thefootnote{}\footnote{#1}%
  \addtocounter{footnote}{-1}%
  \endgroup
}

\ifcvprfinal\fi
\begin{document}

\title{Network Pruning via Resource Reallocation}

\author{\textbf{Yuenan Hou}$^{1}$, \textbf{Zheng Ma}$^{2}$, \textbf{Chunxiao Liu}$^{2}$, \textbf{Zhe Wang}$^{2}$, \textbf{and Chen Change Loy}$^{3\dagger}$\\
$^{1}$The Chinese University of Hong Kong $^{2}$SenseTime Group Limited $^{3}$S-Lab, Nanyang Technological University\\
$^{1}$hy117@ie.cuhk.edu.hk, $^{2}$\{mazheng, liuchunxiao, wangzhe\}@sensetime.com, $^{3}$ccloy@ntu.edu.sg
}

\maketitle

\def\algorithmname{PEEL}

\begin{abstract}
Channel pruning is broadly recognized as an effective approach to obtain a small compact model through eliminating unimportant channels from a large cumbersome network. Contemporary methods typically perform iterative pruning procedure from the original over-parameterized model, which is both tedious and expensive especially when the pruning is aggressive. In this paper, we propose a simple yet effective channel pruning technique, termed network Pruning via rEsource rEalLocation (\algorithmname), to quickly produce a desired slim model with negligible cost. 
Specifically, \algorithmname~first constructs a predefined backbone and then conducts resource reallocation on it to shift parameters from less informative layers to more important layers in one round, thus amplifying the positive effect of these informative layers. To demonstrate the effectiveness of \algorithmname~, we perform extensive experiments on ImageNet with ResNet-18, ResNet-50, MobileNetV2, MobileNetV3-small and EfficientNet-B0. Experimental results show that structures uncovered by \algorithmname~exhibit competitive performance with state-of-the-art pruning algorithms under various pruning settings. Our code is available at \url{https://github.com/cardwing/Codes-for-PEEL}.
\end{abstract}

\section{Introduction}
\label{sec:introduction}

\blfootnote{$\dagger$: Corresponding author.}

Recent years have witnessed the great success of convolution neural networks in many computer vision tasks, \eg, image classification~\cite{he2016deep}, semantic segmentation~\cite{long2015fully} and object detection~\cite{szegedy2013deep}. The remarkable performance usually comes at the cost of dramatically increased network complexity, which makes these models prohibitive for resource-limited edge devices and mobile applications. Channel pruning~\cite{li2017pruning,luo2017thinet} has been acknowledged as an effective algorithm to notably reduce the model's demand on computational and storage resources via discarding less informative channels of the original model.

Conventional channel pruning approaches~\cite{liu2017learning,guo2020dmcp,yu2019universally} typically remove redundant channels from a large, cumbersome neural network to acquire a compact model. For instance, Guo \etal~\cite{guo2020dmcp} introduce additional parameters to learn the retaining probability of each individual channel and prune unnecessary channels based on the learned probability value. Liu \etal~\cite{liu2017learning} add L$_{1}$ regularization of Batch Normalization (BN) statistics to the training process and recursively eliminate less important channels according to the magnitude of BN statistics. These channel pruning techniques either entail extra learnable parameters or involve iterative and expensive pruning procedure, which prohibits many of them from real-world applications. 
 
\begin{figure}[t]
 \centering
 \includegraphics[width=1.0\linewidth]{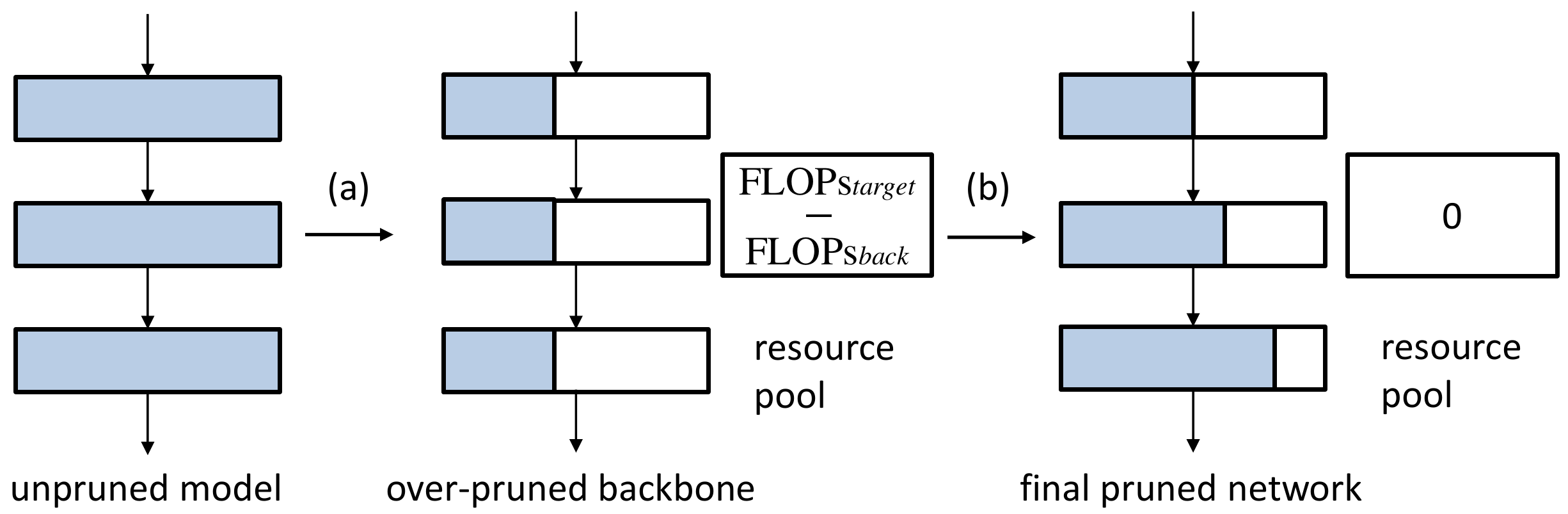}
 \caption{Illustration of \algorithmname. Given the original unpruned model and target $\texttt{FLOPs}_{target}$, \algorithmname~performs the following operations: (a) construct an over-pruned backbone model whose $\texttt{FLOPs}_{back}$ is smaller than the target value $\texttt{FLOPs}_{target}$ and store the extra FLOPs ($\texttt{FLOPs}_{target} - \texttt{FLOPs}_{back}$) in a resource pool, (b) reassign the resources in the pool to different layers of this backbone model based on the estimated layer importance and obtain the desired compact model. The blue part in each block denotes the remaining channels in a layer and the white part represents the pruned channels.}
 \centering
 \vskip -0.2cm
 \label{fig:method_brief_view}
\end{figure}

In this paper, we propose a conceptually simple yet effective channel pruning approach to obtain a compact and good-performing network with negligible cost. Our algorithm is based on the premise that all layers do not contribute equally to the final performance of the network. Layers that have indispensable effect on the network behavior are regarded as important while other layers are considered as less crucial. By shifting resources from less crucial layers to more significant layers, one can easily acquire a desired slim model from off-the-shelf network architectures.

Our method is known as network pruning via resource reallocation (\algorithmname). As the name implies, \algorithmname~addresses channel pruning via reallocating the resources on a well-established backbone model. As shown in Fig.~\ref{fig:method_brief_view}, we first construct an over-pruned backbone model and store the saved resources (\eg, FLOPs, parameters or latency) in a resource pool. By estimating layer importance via some standard criteria, resource reallocation is then performed to assign resources to different layers based on the estimated layer significance in one round.

\algorithmname~offers the possibility of reassigning parameters on a backbone architecture. As opposed to previous backward elimination methods, \algorithmname~is built upon forward selection on a predefined architecture. It does not introduce additional learnable parameters and is free from expensive iterative pruning procedure. Despite its simplicity, \algorithmname~can find good structures based on diverse backbone models on ImageNet, and its performance is on par with the architectures pruned by state-of-the-art channel pruning algorithms (\eg, DMCP~\cite{guo2020dmcp}) under various pruning levels.

In summary, we contribute a simple yet effective channel pruning algorithm, \algorithmname, to produce a desired compact model via performing resource reallocation on a pre-defined backbone model. Through extensive experiments, we verify that \algorithmname~can produce stable searching results with small performance variance. Besides, it is applicable to various backbone models and is robust to the criterion to determine the layer importance. We validate the effectiveness of \algorithmname~on modern CNN architectures, \ie, ResNet-18, ResNet-50, MobileNetV2, MobileNetV3-small and EfficientNet-B0, on the large-scale ImageNet benchmark. Under diverse FLOPs constraints, \algorithmname~can consistently find targeted slim models with less searching cost compared with the state-of-the-art channel pruning techniques, \eg, DMCP~\cite{guo2020dmcp}, FPGM~\cite{he2019filter} and USNet~\cite{yu2019universally}. We hope \algorithmname~can serve as a strong baseline to facilitate future research on channel pruning. 

\section{Related Work}
\label{sec:relatedwork}

\noindent \textbf{Channel Pruning:} Channel pruning is a prevailing approach to remove redundant channels from the over-parameterized neural networks. Typically, channel pruning is composed of three stages, \ie, training the original network, pruning unimportant channels and finetuning the pruned architecture~\cite{liu2017learning,han2015learning}. 

According to the pruning directions, contemporary channel pruning algorithms can be categorized into two types, \ie, backward elimination from the original model~\cite{guo2020dmcp,gao2018pruning,han2015learning,he2018amc,he2019filter,liu2017learning,yu2019autoslim} and forward selection from an empty network~\cite{ye2020good,zhuang2018discrimination}. The majority of the pruning methods fall into the former type and they delete unnecessary connections based on diverse pruning criteria, \eg, the magnitude of kernel weights~\cite{han2015learning} or BN statistics~\cite{liu2017learning}, the feature map reconstruction errors~\cite{luo2017thinet} or the absolute first-order gradient of objective function~\cite{molchanov2017pruning}. For instance, Liu \etal~\cite{liu2017learning} add L$_{1}$ regularization to the network training phase to enforce network sparsity and remove channels whose BN statistic is smaller than a preset threshold. Guo \etal~\cite{guo2020dmcp} phrase channel pruning as a Markov process and introduce additional parameters to each channel to learn the probability of pruning. Contrary to the preceding indirect pruning algorithms, another line of work~\cite{yu2019autoslim,yu2019universally} concentrates on training a supernet and directly takes the accuracy of the sampled structures as the measurement. For example, Yu \etal ~\cite{yu2019autoslim} first train a universally slimmable super-network and then exhaustively evaluate a collection of sampled sub-networks from the trained supernet to determine the optimal substructure. Nonetheless, the aforementioned work suffers from expensive training and searching cost especially when the pruning is aggressive since either iterative pruning procedure or many rounds of evaluation are required to find the best-performing architecture. 

As to the forward selection, Ye \etal ~\cite{ye2020good} start from an empty network and gradually add important neurons from the original model in the greedy manner. Zhuang \etal ~\cite{zhuang2018discrimination} recursively place important channels in each empty layer based on the gradient norm. Our algorithm belongs to the forward selection category but has several distinctions with ~\cite{ye2020good,zhuang2018discrimination}. First, we start from a predefined backbone instead of an empty architecture therefore making our selection process more efficient. Second, in contrast to previous efforts that add informative neurons one by one, \algorithmname~first estimates the importance of different layers and then reallocate resources to these layers via the estimated importance in one round, which substantially relieves the selection burden.

Our work is related to network rejuvenation~\cite{qiao2019neural}. However, the ultimate objective of~\cite{qiao2019neural} is to increase the resource utilization ratio and it is not dedicated to the network pruning task. In all, our \algorithmname~differs from their method in four aspects. First, \algorithmname~reassigns resources according to the computed layer importance while \cite{qiao2019neural} allocates the resources based on the number of remaining channels in each layer. The allocation policy of \cite{qiao2019neural} heavily hinges on the quality of the backbone model and may suffer a lot if the architecture of the backbone model is irrational. Second, \cite{qiao2019neural} requires sophisticated procedure, \eg, parameter reinitialization, neural rescaling and well-designed training scheme, to train the resulting model while \algorithmname~just entails the vanilla training strategy. Third, \cite{qiao2019neural} has several hyperparameters to tune, \eg, the sparsity coefficient, the pruning threshold for each layer and the utilization threshold to conduct neural rejuvenation, while \algorithmname~has only one hyperparameter. Fourth, \cite{qiao2019neural} conducts experiments only on mild pruning levels on ImageNet whereas \algorithmname~includes mild, medium and aggressive pruning settings.    

\noindent \textbf{Knowledge Distillation:} Knowledge distillation~\cite{hinton2015distilling} is widely acknowledged as an effective technique to transfer the dark knowledge from the large over-parameterized network to the small compact model. The large network is called teacher and the small model is dubbed student. The transferred knowledge can take on many forms, \eg, softened output logits~\cite{hinton2015distilling}, visual attention maps~\cite{zagoruyko2016paying,hou2020inter} or intermediate feature maps~\cite{liu2019structured,hou2019learningto}. The above distillation algorithms typically consume huge memory storage since they need to simultaneously load both student and teacher during training. To circumvent the participation of cumbersome teacher models, recent studies~\cite{sun2019deeply,hou2019learning,zhang2019your} have focused on deriving the supervision signals from the student model itself. For instance, Hou \etal reinforce the representation learning of shallow layers via attention maps of deep layers~\cite{hou2019learning}. Our algorithm only adopts the simplest form of knowledge distillation~\cite{hinton2015distilling} to compare fairly with other pruning algorithms. Nevertheless, the aforementioned distillation approaches can also be employed to further boost the performance of the final compact model.

\section{Methodology}
\label{sec:methodology}

Channel pruning can be formulated as follows: given the target resource constraint (\eg, FLOPs, parameters, latency), the objective is to find a set of channel configurations that achieves the optimal performance on the target task. Considering that the possible number of channels in each layer is a non-negative integer, we phrase channel pruning as a combinatorial optimization problem. The specific mathematical formulation is presented below:
\begin{equation}
\begin{aligned}
\max_{(c_{1},...,c_{N})} \quad & f(c_{1},...,c_{N}).\\
\textrm{s.t.} \quad & e(c_{1},...,c_{N}) \leq M\\
  &0\leq c_{i} \leq C_{i}    \\
  &0\leq i \leq N\\
\end{aligned}
\end{equation}
Here, $f$ is the performance estimation function of the model, $e$ is the resource estimation function, $M$ is the target resource contraint (\eg, $\texttt{FLOPs}_{target}$ in Fig.~\ref{fig:method_brief_view}), $c_{i}$ and $C_{i}$ are the channel number in the $i$-th layer of the pruned and original unpruned model, respectively, $N$ is the number of layers in the unpruned network. 

Conventional approaches~\cite{yu2019universally,guo2020dmcp,liu2019metapruning,liu2017learning} typically discard unnecessary channels from the original over-parameterized network. Distinct from previous backward elimination methods, the proposed \algorithmname~addresses channel pruning via reallocating parameters in a predefined architecture, usually an over-pruned backbone from the original network. The resource consumption of this predefined structure is smaller than the target value and the remaining resources are stored in a place named resource pool (Fig.~\ref{fig:method_brief_view}). The intuition of \algorithmname~is that those informative layers should be assigned more parameters to amplify their positive effect on the final performance.

The proposed \algorithmname~is comprised of five steps, namely, the construction of the over-pruned backbone and resource pool, estimation of layer importance, layer grouping, reallocation of parameters to the backbone based on the estimated importance, and finally, the retraining the resulting compact model with knowledge distillation.

\noindent \textbf{Step 1 - Construction of the over-pruned backbone and resource pool:} Given the target resource constraints $M$, we first build an over-pruned backbone model. In this paper, we adopt uniform pruning~\cite{liu2018rethinking} to construct the backbone model. Uniform pruning possesses many attractive properties. First, it preserves the original architecture as much as possible and therefore the pruned network shares more structural similarity with the original model. Such structural resemblance will facilitate the knowledge distillation process and can bring more gains to the pruned model. Second, contrary to previous pruning methods that require the training of the original unpruned network, uniform pruning introduces no training and searching cost since the pruning ratio is same for all layers. Third, when the pruning is mild, \eg, pruning fewer than 40 $\%$ channels, uniform pruning can serve as a strong baseline, with its performance comparable with that of contemporary pruning algorithms. Consequently, we employ uniform pruning to construct a simple backbone model. 

We next prepare the resource pool.
We define the proportion of the resources the backbone model consumes (\eg, $\texttt{FLOPs}_{back}$ in Fig.~\ref{fig:method_brief_view}) to the given target resource constraint $M$ as $\lambda$ ($0\leq\lambda\leq1$), which is a tunable hyperparameter in our algorithm.
The resource pool thus stores the extra resources of $(1-\lambda)M$, \eg, $\texttt{FLOPs}_{target} - \texttt{FLOPs}_{back}$ in Fig.~\ref{fig:method_brief_view}.
Intuitively, a high $\lambda$ suggests few resources will be kept in the resource pool. 
With a small $\lambda$, more resources will be stored in the pool by having a much slimmer over-pruned backbone.
In our experiments, $\lambda$ is empirically set as 0.8 since this value yields satisfactory results. We avoid an overly small $\lambda$ observing that over-aggressive pruning will hurt the backbone's performance, which could hardly be compensated by subsequent operations. It is noteworthy that although we leverage uniform pruning to establish the backbone model, \algorithmname~can still work well when taking different models as backbones.

\noindent \textbf{Step 2 - Estimation of layer importance:} One crucial drawback of uniform pruning is that it considers all layers to be of equal importance and forces the same pruning ratio for all layers. This practice neglects the fact that some layers contribute more than other layers to the model performance, therefore these influential layers should be assigned more parameters. Hence, one core objective of \algorithmname~is to distinguish important layers from less informative layers and place more parameters in these informative layers. 

There are several options for the evaluating criteria on estimating layer importance, \eg, weight magnitude~\cite{han2015learning}, reconstruction errors of layer activations~\cite{luo2017thinet}, BN statistics~\cite{liu2017learning}, and gradient of the cost function~\cite{molchanov2017pruning}. We eventually choose the BN statistics, \ie, absolute gamma value, as the criterion to reflect the importance of each layer to the final performance as it yields appealing results and requires little computation cost. Batch normalization~\cite{ioffe2015batch} has been a basic block in modern CNN architectures to address the internal covariate shift. Suppose $X_{i}$ and $X_{o}$ are the input and output of the BN layer, respectively. BN layer performs the following transformation on the input:
\begin{equation}
\label{eqn:bn}
\begin{split}
& X_{o} = \gamma \frac{X_{i} - \mu}{\sqrt{\sigma^{2} + \epsilon}} + \beta,
\end{split}
\end{equation}
where $\mu$ and $\sigma$ are the mean and standard deviation of the current input along the channel dimension, respectively, $\epsilon$ is a small value to ensure numerical stability, $\gamma$ and $\beta$ are two learnable parameters for the BN layer. Following~\cite{liu2017learning}, we add the L$_{1}$-norm of $\gamma$ to the main loss to encourage sparsity:
\begin{equation}
\label{eqn:search_loss}
\begin{split}
& \mathcal{L}_{train} = \mathcal{L}_\mathrm{cls}(\mathbf{O}_{S}, \mathbf{O}_{L}) + \alpha_{s} \sum\nolimits_{i=1}^{N_{\gamma}} |\gamma_{i}|,
\end{split}
\end{equation}
where $N_{\gamma}$ is the number of all $\gamma$ in the network, $\alpha_{s}$ is the sparsity coefficient, $\mathbf{O}_{S}$ is the model prediction, $\mathbf{O}_{L}$ is the ground-truth label and $\mathcal{L}_\mathrm{cls}$ is the cross entropy loss.

\begin{figure}[t]
 \centering
 \includegraphics[width=0.8\linewidth]{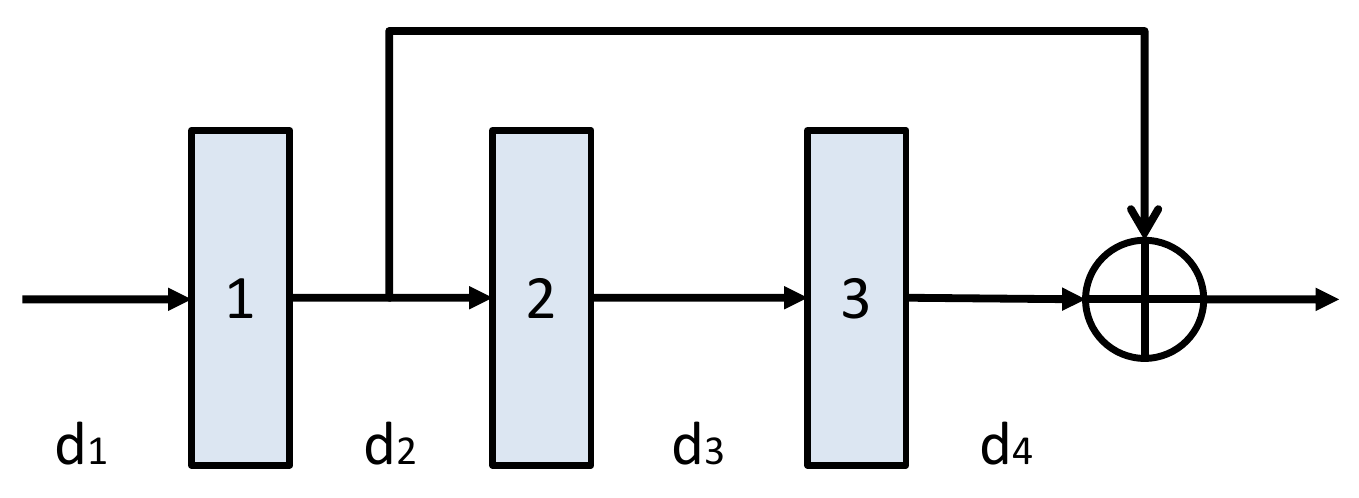}
 \vskip -0.1cm
 \caption{Illustration of pruning channels in residual blocks. For layer $i$, the number of input and output channels are $d_{i}$ and $d_{i+1}$, respectively. The output of layer 1 and 3 are added element-wise via the residual connection. Hence, $d_{2}$ and $d_{4}$ have to be kept equal after pruning.}
 \centering
 \vskip -0.2cm
 \label{fig:residual_connect}
\end{figure}

\noindent \textbf{Step 3 - Layer grouping:} 
Considering the fact that modern CNN architectures, \eg, ResNet, have residual connections, pruning channels in these networks is troublesome since the output of one layer is directly added to that of another layer, and the channel dimension of these layers has to be kept the same. An example is presented in Fig.~\ref{fig:residual_connect}. To prune these three layers, we have to keep $d_{2}$ equal to $d_{4}$. Moreover, ResNet typically has many residual blocks, thus posing more constraints on the consistency in the channel dimensions of multiple layers. To address the aforementioned issue, we decide to group convolutional layers according to their positions in the original network. More concretely, layers whose feature maps have the same spatial size are grouped together. Under this condition, the output channels of the layers in one group will undergo the same linear expansion or shrinkage, which satisfies the dimension consistency requirement raised by residual connections.

\noindent \textbf{Step 4 - Resource reallocation:} After the layer grouping process, we need to properly assign the remaining resources in the resource pool to those groups of layers. There are two options for the resource reallocation procedure, \ie, allocating the additional parameter resources in one round or in multiple rounds. The primary distinction between these two options is that the former only considers layer importance in the current uniformly pruned model while the latter takes into account the ever-changing layer significance in various intermediate models. Apparently, the one-round practice will consume much less computation budget than the multiple-round version. Here, we focus on the one-round allocation as it can already bring appealing results and leave the comparison between one-round and multi-round assignment in the ablation study.

Suppose there are $G$ groups in total after layer grouping and the total number of their output channels in group $i$ is $g_{i}$. Then, for group $i$, its importance is calculated as:
\begin{equation}
\label{eqn:group_importance}
\begin{split}
T_{i} = \frac{1}{g_{i}} \sum\nolimits_{j=1}^{g_{i}} |\gamma_{j}|.
\end{split}
\end{equation}
Then, the resources assigned to group $i$ are:
\begin{equation}
\label{eqn:group_resource}
\begin{split}
R_{i} = \frac{T_{i}}{\sum_{j=1}^{G} T_{j}}(1-\lambda)M,
\end{split}
\end{equation}
where $(1-\lambda)M$ is the resources available in the resource pool.
In each group, we adopt the linear expansion strategy, \ie, the channels of all layers in one group are multiplied by a common value. In other words, the increased channels of one layer is proportional to its original number of channels. After performing the linear expansion on each layer, we eventually obtain the desired compact model.

\noindent \textbf{Step 5 - Distilling knowledge from the original model:} Once the compact model is obtained after resource allocation, we follow the common practice in network pruning~\cite{guo2020dmcp,yu2019universally} and start to train this model from scratch with both hard label and soft label supervision signals. The hard label denotes the ground-truth one-hot vector and the soft label represents the softened probability distribution from the original cumbersome model~\cite{hinton2015distilling}. By distilling the probabilistic knowledge from the trained unpruned network, the small compact model can learn better representations and exhibit better generalization capability after the training phase. The final loss function is written as below, which is comprised of the hard label loss and the distillation loss:
\begin{equation}
\label{eqn:total_loss}
\begin{split}
& \mathcal{L}_{retrain} = \mathcal{L}_\mathrm{cls}(\mathbf{O}_{S}, \mathbf{O}_{L}) + \alpha_{d} \mathcal{L}_\mathrm{distill}(\textbf{P}_{S}, \textbf{P}_{T}).
\end{split}
\end{equation}
Here, $\mathbf{O}_{S}$ is the probability output of the compact model, $\mathbf{O}_{L}$ is the ground-truth label, $\mathbf{P}_{S}$ and $\mathbf{P}_{T}$ are the output probabilities of the pruned and unpruned model, respectively. We denote $\alpha_{d}$ as the coefficient to balance the hard label loss and distillation loss, $\mathcal{L}_\mathrm{cls}$ is the cross entropy loss and $\mathcal{L}_\mathrm{distill}$ is the KL divergence loss. Note that $P_{S}$ and $P_{T}$ are obtained via performing softmax on the output logits~\cite{hinton2015distilling} and the temperature is set as 1.

\section{Experiments}
\label{sec:experiments}

Following~\cite{guo2020dmcp}, we perform extensive experiments on ImageNet under various pruning levels. There are 1000 classes in ImageNet, around 128M images are used for training and 50K images for validation. We choose modern CNN architectures, \ie, ResNet-18~\cite{he2016deep} and ResNet-50~\cite{he2016deep}, for pruning as they are widely used in both academia and industry. Following~\cite{guo2020dmcp}, we also perform pruning on MobileNetV2~\cite{sandler2018mobilenetv2}. This is a more challenging task since MobileNetV2 is already a compact model. We also report results on pruning extremely efficient architectures, \ie, MobileNetV3-small~\cite{Howard_2019_ICCV} and EfficientNet-B0~\cite{tan2019efficientnet}. It is noteworthy that \algorithmname~is not limited to these architectures and can be employed to address the pruning of more sophisticated networks.  

\noindent \textbf{Evaluation metric:} Following~\cite{guo2020dmcp}, we adopt the top-1 accuracy on the center crop as the evaluation metric.

\noindent \textbf{Implementation details:} To ensure a fair comparison, we adopt the exact setting of~\cite{guo2020dmcp}. Concretely, the total number of epochs for training the uniformly pruned backbone is 50. The final pruned architecture is trained for 100 epochs for ResNets, 250 epochs for MobileNetV2, and the first epoch is used to warm up the pruned model. Batch size is set as 512 and the initial learning rate is set as 0.2. The learning rate is gradually reduced to zero via a cosine learning rate schedule. Label smoothing is employed to relieve the over-fitting issue, and the corresponding smoothing factor is set as 0.1. Parameters $\alpha_{s}$ and $\alpha_{d}$ are set as 0.0001 and 0.1, respectively. As to data augmentation, random cropping and resizing, random horizontal flipping, color jitter and normalization are orderly applied to the input images during training. For knowledge distillation, ResNet-18 and ResNet-50 are trained with the original distillation strategy~\cite{hinton2015distilling}. For MobileNetV2, we follow~\cite{guo2020dmcp,yu2019autoslim} and make use of the in-place distillation to train the compact model.   

\noindent \textbf{Baseline algorithms:} DMCP~\cite{guo2020dmcp} is currently the most competitive algorithm in channel pruning. It models channel pruning as a Markov Decision Process, where the state is the remaining channels in a layer, and the transition from one state to another state represents the pruning process. By fusing the learnable transition probabilities with the layer output, the whole system is optimized end-to-end with gradient descent and can be easily combined with specific FLOPs constraint to acquire the desired network architecture. The training cost of DMCP is composed of two parts, \ie, the learning of pruning probabilities and the training of the pruned model. The sampling procedure to acquire the compact model consumes negligible resources since they employ expected sampling to obtain the desired network in one round. In the subsequent paragraphs, we will systematically compare our \algorithmname~with DMCP in terms of the overall performance under diverse FLOPs requirement, the expected training cost, the variability of the searched structures as well as performance stability. FPGM~\cite{he2019filter}, NR~\cite{qiao2019neural}, MetaPruning~\cite{liu2019metapruning}, AutoSlim~\cite{yu2019autoslim} and AMC~\cite{he2018amc} are also included in the performance comparison since there are all contemporary channel pruning techniques.   

\subsection{Results}

\begin{table}
\caption{Performance of \algorithmname~and various channel pruning approaches with and without knowledge distillation (KD) on ImageNet validation set.}
\label{result_table}
\centering
\footnotesize{
\begin{tabular}{c|c|c|c|c}
\hline
\multirow{2}*{Backbone} & \multirow{2}*{Method} & \multicolumn{2}{|c|}{Top-1 Acc (\%)} & \multirow{2}*{FLOPs} \\
\cline{3-4}
~ & ~ & w/ KD & w/o KD & ~ \\
\hline 
\multirow{10}*{ResNet-18} & Uniform 1.0 $\times$ & \multicolumn{2}{|c|}{70.3} & 1.8G \\
\cline{2-5}
~ & Uniform 0.85 $\times$ & 69.0 & 68.5 & 1.33G \\
~ & NR~\cite{qiao2019neural} & 68.6 & 68.2 & 1.33G \\
~ & DMCP~\cite{guo2020dmcp} & 70.0 & 69.7 & 1.33G \\
~ & \textbf{\algorithmname} & \textbf{70.6} & 70.1 & 1.33G \\
\cline{2-5}
~ & Uniform 0.75 $\times$ & 68.2 & 67.8 & 1.05G \\
~ & NR~\cite{qiao2019neural} & 68.1 & 67.9 & 1.04G \\
~ & FPGM~\cite{he2019filter} & 68.4 & 68.1 & 1.04G  \\
~ & DMCP~\cite{guo2020dmcp} & 69.4 & 69.0 & 1.04G \\
~ & \textbf{\algorithmname} & \textbf{69.9} & 69.3 & 1.04G \\
\hline
\multirow{18}*{ResNet-50} & Uniform 1.0 $\times$ & \multicolumn{2}{|c|}{76.4} & 4.1G \\
\cline{2-5}
~ & Uniform 0.85 $\times$ & 75.8 & 75.4 & 3.0G \\
~ & NR~\cite{qiao2019neural} & 75.2 & 74.8 & 3.0G \\
~ & MetaPruning~\cite{liu2019metapruning} & 76.2 & 76.0 & 3.0G \\
~ & AutoSlim~\cite{yu2019autoslim} & 76.0 & 75.8 & 3.0G \\
~ & DMCP~\cite{guo2020dmcp} & 76.5 & 76.4 & 3.0G \\
~ & \textbf{\algorithmname} & \textbf{76.9} & 76.6 & 3.0G \\
\cline{2-5}
~ & Uniform 0.75 $\times$ & 74.6 & 74.1 & 2.3G \\
~ & NR~\cite{qiao2019neural} & 74.3 & 74.0 & 2.4G \\
~ & MetaPruning~\cite{liu2019metapruning} & 75.4 & 75.2 & 2.3G \\
~ & FPGM~\cite{he2019filter} & 75.6 & 75.5 & 2.4G \\
~ & DMCP~\cite{guo2020dmcp} & 76.3 & 76.2 & 2.2G \\
~ & \textbf{\algorithmname} & \textbf{76.8} & 76.5 & 2.2G \\
\cline{2-5}
~ & Uniform 0.5 $\times$ & 72.9 & 72.7 & 1.1G \\
~ & NR~\cite{qiao2019neural} & 73.1 & 72.6 & 1.1G \\
~ & MetaPruning~\cite{liu2019metapruning} & 73.4 & 73.2 & 1.1G \\
~ & AutoSlim~\cite{yu2019autoslim} & 74.0 & 73.6 & 1.1G \\
~ & DMCP~\cite{guo2020dmcp} & 74.6 & 74.4 & 1.1G \\
~ & \textbf{\algorithmname} & \textbf{75.1} & 74.7 & 1.1G \\
\hline
\multirow{21}*{MobileNetV2} & Uniform 1.0$\times$ & \multicolumn{2}{|c|}{72.3} & 300M \\
\cline{2-5}
~ & NR~\cite{qiao2019neural} & 73.0 & 72.2 & 300M \\
~ & AutoSlim~\cite{yu2019autoslim} & 74.2 & 73.1 & 300M \\
~ & DMCP~\cite{guo2020dmcp} & 74.6 & 73.9 & 300M \\
~ & \textbf{\algorithmname} & \textbf{74.8} & 74.2 & 300M \\
\cline{2-5}
~ & Uniform 0.75 $\times$ & 70.1 & 69.2 & 210M \\
~ & NR~\cite{qiao2019neural} & 70.2 & 69.4 & 211M \\
~ & MetaPruning~\cite{liu2019metapruning} & 71.2 & 70.1 & 217M \\
~ & AMC~\cite{he2018amc} & 70.8 & 70.0 & 211M \\
~ & AutoSlim~\cite{yu2019autoslim} & 73.0 & 72.2 & 211M \\
~ & DMCP~\cite{guo2020dmcp} & 73.5 & 72.4 & 211M \\
~ & \textbf{\algorithmname} & \textbf{73.9} & 73.0 & 211M \\
\cline{2-5}
~ & Uniform 0.5 $\times$ & 64.8 & 64.3 & 97M \\
~ & NR~\cite{qiao2019neural} & 64.2 & 63.8 & 87M \\
~ & MetaPruning~\cite{liu2019metapruning} & 63.8 & 63.4 & 87M \\
~ & DMCP~\cite{guo2020dmcp} & 66.1 & 65.6 & 87M \\
~ & \textbf{\algorithmname} & \textbf{66.6} & 66.0 & 87M \\
\cline{2-5}
~ & Uniform 0.35$\times$ & 60.1 & 59.7 & 59M \\
~ & NR~\cite{qiao2019neural} & 59.7 & 59.2 & 59M \\
~ & DMCP~\cite{guo2020dmcp} & 62.7 & 62.4 & 59M \\
~ & \textbf{\algorithmname} & \textbf{62.9} & 62.6 & 59M \\
\cline{2-5}

\hline
\end{tabular}
}
\end{table}

\noindent\textbf{Comparisons with state of the arts:}
The detailed performance comparison of \algorithmname~and contemporary channel pruning algorithms is shown in Table~\ref{result_table}. We also record the floating-point operations (FLOPs) to approximate the forward time of different architectures. From Table.~\ref{result_table}, we observe that \algorithmname~achieves 25$\%$ and 45$\%$ FLOPs reduction for ResNet-18 and ResNet-50, respectively, with almost no loss of accuracy. 
As to MobileNetV2, the result is more encouraging as our algorithm can find a strong architecture with \textbf{2.5\%} higher in top-1 accuracy than the original model while possessing the same FLOPs. In addition, \algorithmname~consistently finds better architectures whose performance outperforms those searched by other competitive channel pruning techniques, \eg, DMCP~\cite{guo2020dmcp}, NR~\cite{qiao2019neural} and MetaPruning~\cite{liu2019metapruning}. For instance, when the backbone model is ResNet-50 and the target FLOPs is 2.2G, the architecture produced by \algorithmname~is \textbf{0.5\%} higher in top-1 accuracy than that found by DMCP. Moreover, the network structure obtained via \algorithmname~consistently outperforms the uniformly pruned model in all experiments, suggesting the effectiveness of the proposed resource reallocation approach.

\begin{figure}[!t]
 \centering
 \includegraphics[width=1.0\linewidth]{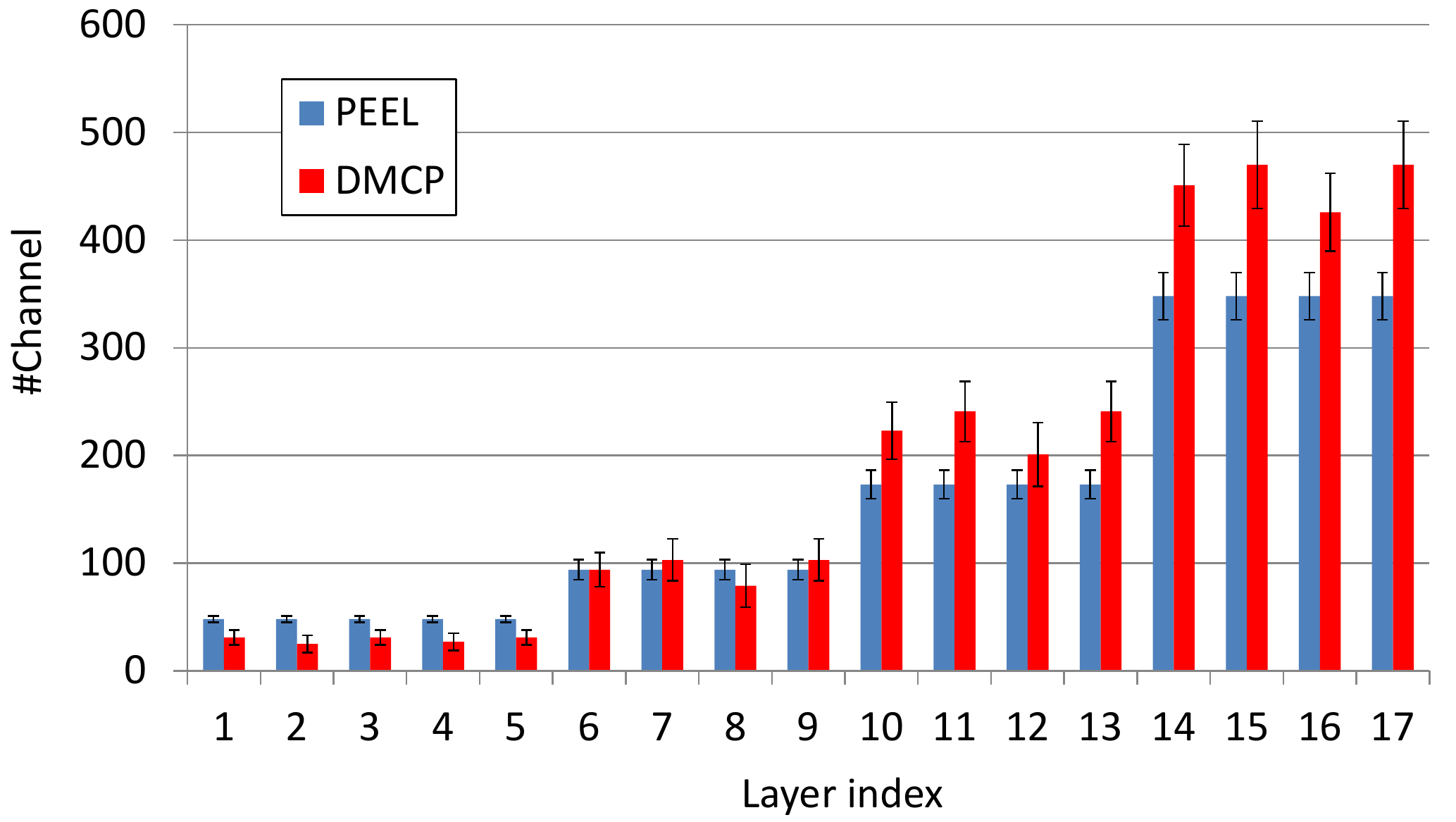}
 \vskip -0.1cm
 \caption{Comparison between structures searched by \algorithmname~and DMCP on ResNet-18 when $\#$FLOPs=1.04G. The vertical bar and vertical line denote the mean and variance of the channel numbers in each layer, respectively. Results are averaged over five runs.}
 \centering
 \label{fig:arch_compare}
\end{figure}

\begin{table}[!t]
\caption{Pruning MobileNetV3-small and EfficientNet-B0.}
\label{prune_harder_model_table}
\centering
\small{
\begin{tabular}{c|c|c|c}
\hline
Backbone & Method & Top-1 Acc (\%) & FLOPs \\
\hline
\multirow{2}*{MobileNetV3} & Uniform 1.0 $\times$ & 67.4 & 56M \\
\cline{2-4}
~ & \algorithmname & 67.2 & 49M \\
\hline
\multirow{2}*{EfficientNet-B0} & Uniform 1.0 $\times$ & 77.1 & 390M \\
\cline{2-4}
~ & \algorithmname & 77.0 & 346M \\
\hline
\end{tabular}
}
\vspace{-5ex}
\end{table}

\noindent \textbf{Pruning MobileNetV3-small and EfficientNet-B0:} As shown in Table~\ref{prune_harder_model_table}, \algorithmname~saves \textbf{12.5\%} FLOPs for MobileNetV3-small and \textbf{11.3\%} FLOPs for EfficientNet-B0 without incurring severe performance drops. Note that MobileNetV3-small and EfficientNet-B0 are two extremely lightweight classification models and removing redundancy in these models is nontrivial. The experimental results explicitly showcase the effectiveness and generality of \algorithmname~on network pruning.

We also visualize the structures uncovered by \algorithmname~and DMCP. As depicted in Fig.~\ref{fig:arch_compare}, both architectures have more channels as the layer goes deeper. This is expected since more channels will be leveraged to compensate for the loss of spatial information incurred by the downsampling operations. Nevertheless, \algorithmname~tends to place more channels in the early stages while DMCP chooses to put more channels in the later stages. Putting more resources at shallow layers brings more gains as there is more redundancy in the deep layers. Besides, the variance of the number of channels in each layer of \algorithmname~is much smaller than that of DMCP. For example, the averaged channel variance of DMCP on the last four layers of ResNet-18 is \textbf{38.8} while the channel variance of \algorithmname~is \textbf{21.8}. The gap between the structural variance of DMCP and our algorithm is more evident in ResNet-50 and MobileNetV2. In ResNet-50, the averaged channel variance of \algorithmname~is almost half of the variance of DMCP for all layers. We provide more details in the supplementary material.
The searching instability of DMCP also leads to high performance variance in the searched architectures. The performance of PEEL is  much stabler relatively. We summarize the performance of architectures searched by \algorithmname~and DMCP in Table~\ref{performance_var_table}. 


\begin{table}[t]
\caption{Performance variance of searched architectures of PEEL and DMCP on ResNet-50. Results are averaged over five runs.}
\vskip 0.1cm
\label{performance_var_table}
\centering
\small{
\begin{tabular}{c|c|c|c}
\hline
\multirow{2}*{Method} & \multicolumn{3}{c}{$\#$FLOPs} \\
\cline{2-4}
~ & 2.2G & 1.8G & 1.1G \\
\hline
\hline
\textbf{\algorithmname~} & \textbf{76.7\% $\pm$ 0.2\%} & \textbf{75.3\% $\pm$ 0.3\%} & \textbf{75.0\% $\pm$ 0.2\%} \\
\hline
DMCP & 75.7\% $\pm$ 0.7\% & 74.2\% $\pm$ 0.8\% & 73.8\% $\pm$ 0.8\% \\
\hline
\end{tabular}
}
\vspace{-3ex}
\end{table}

\begin{table}[!t]
\caption{Comparison between the training time and memory usage of \algorithmname~and DMCP on ResNet-50 ($\#$FLOPs=1.1G). Here, the memory usage is measured on one GPU and the batch size is set as 64 for each GPU.}
\vskip 0.1cm
\label{train_cost_table}
\centering
\small{
\begin{tabular}{c|c|c}
\hline
Algorithm & Train Time (H) & Memory usage (G) \\
\hline
\hline
\textbf{\algorithmname~} & \textbf{37} & \textbf{4.3} \\
\hline
DMCP~\cite{guo2020dmcp} & 98 & 9.5 \\
\hline
\end{tabular}
}
\vspace{-5ex}
\end{table}

The proposed \algorithmname~not only takes shorter training durations to find and train the desired model, but also consumes less GPU memory during the searching phase. 
Here, we examine the training cost of PEEL and DMCP, \ie, the total training hours and the GPU memory usage during the searching phase. The original unpruned model is ResNet-50 and the target FLOPs is 1.1 G. We keep all other hyperparameters the same, including batch size (512), GPU types (NVIDIA TITAN X) and number of used GPUs (8). From Table~\ref{train_cost_table}, the total number of training hours of \algorithmname~is roughly a third of DMCP. The result is not surprising as DMCP requires training on the original cumbersome model whilst \algorithmname~merely entails training of a much slimmer model. As to the GPU memory usage, since DMCP needs to collect gradients of several sampled architectures while \algorithmname~only computes the gradient of one compact architecture, the memory consumption of DMCP is almost twice as that of \algorithmname~. We also provide a comparison to another representative USNet~\cite{yu2019universally} in the supplementary material.

\begin{figure}[t]
 \centering
 \includegraphics[width=1.0\linewidth]{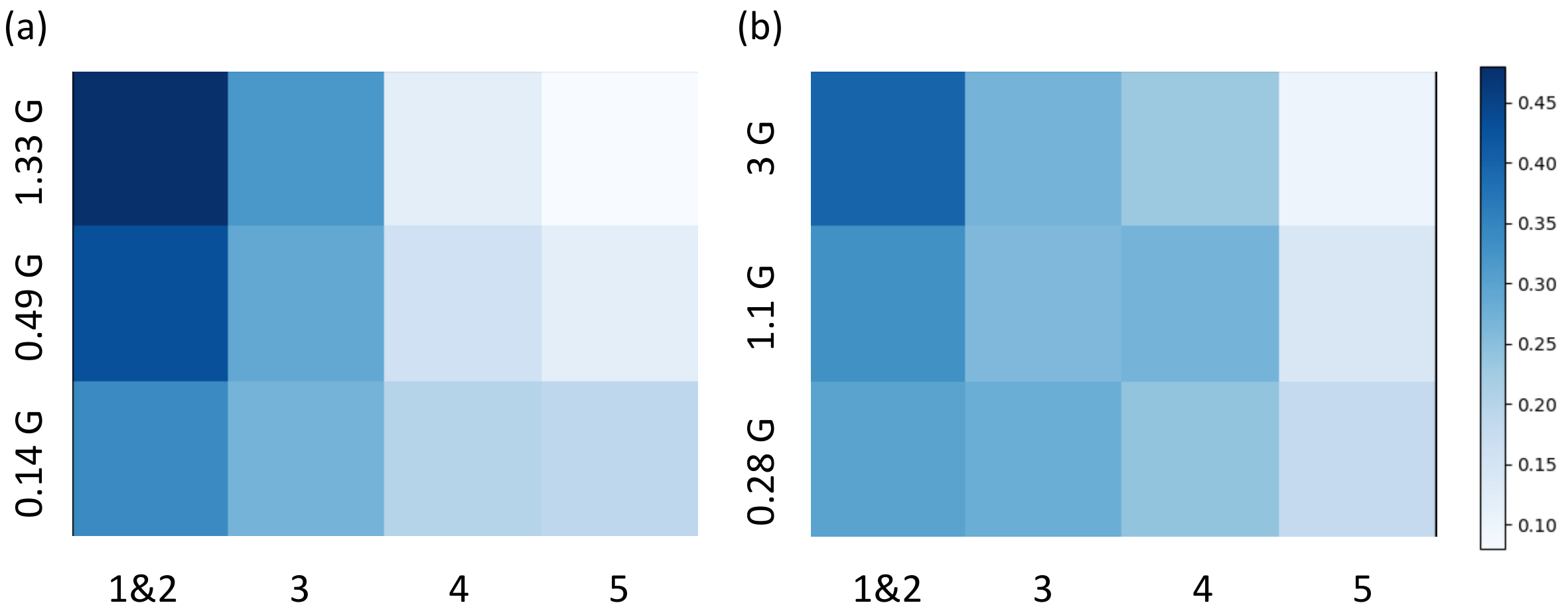}
 \vskip -0.1cm
 \caption{Where the resources go. Percentage of FLOPs assigned by \algorithmname~to different groups of (a) ResNet-18 and (b) ResNet-50 under different target FLOPs constraints. X-axis denotes group indices and y-axis represents the target FLOPs. A darker color indicates a larger quantity of assigned resources. Since ResNet-18 and ResNet-50 have only one layer in group 1 and resources assigned to one layer are very limited, we put this layer into group 2 and obtain the above figure.}
 \centering
 \vskip -0.4cm
 \label{fig:flops_assign_res18}
\end{figure}

\subsection{Ablation study}


\noindent \textbf{Where the resources go:}
To have a deeper understanding on the effect of the resource reallocation module, we visualize the percentage of resources distributed by \algorithmname~to different groups of layers on ResNet-18 and ResNet-50. As shown in Fig.~\ref{fig:flops_assign_res18}, layers in the first three groups are given much more resources than the last two groups when the network is slightly trimmed. It is natural as there are more redundant channels in the deep layers in the mild pruning level~\cite{liu2019metapruning}. As the pruning becomes more aggressive, the percentages of FLOPs assigned to different groups become more even since all layers do not have sufficient channels and call for more resources from the resource pool. And we can observe the same patterns in the resource reallocation of ResNet-18 and ResNet-50 when the FLOPs budget becomes tighter. 

\begin{figure}[t]
 \centering
 \includegraphics[width=1.0\linewidth]{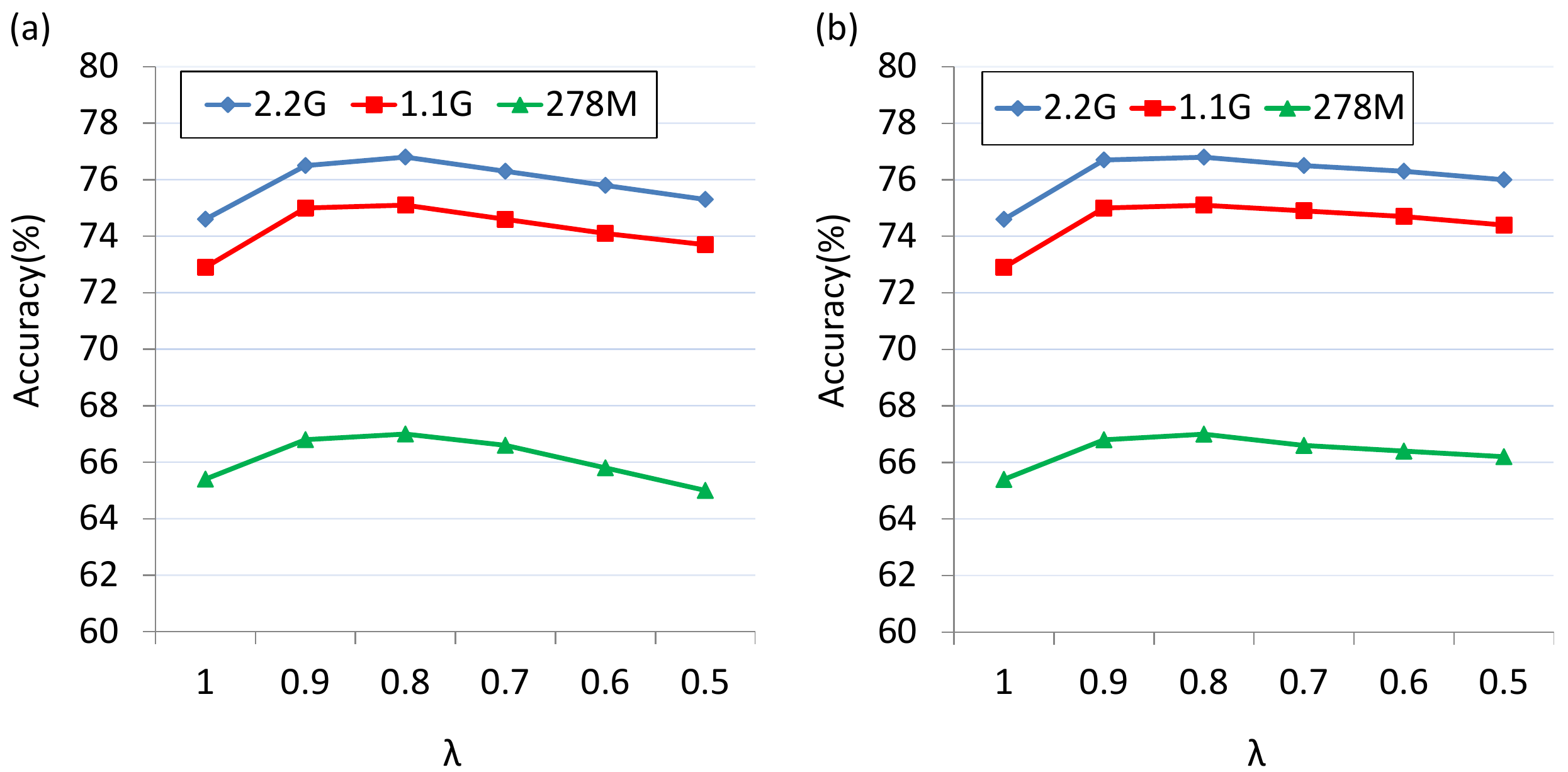}
 \vskip -0.1cm
 \caption{One-round v.s. multi-round reallocation. We show the $\lambda$-accuracy spectrum of (a) one-round and (b) three-round resource reallocation of \algorithmname~on ResNet-50. Different color denotes different target FLOPs.}
 \centering
 \vskip -0.2cm
 \label{fig:single_multi_round}
\end{figure}

\noindent \textbf{Effect of $\lambda$:} The resources available in the resource pool is $(1-\lambda)M$, thus the pool size is controlled by the hyperparameter $\lambda$. We select the value of $\lambda$ from $\{0.5, 0.6, 0.7, 0.8, 0.9, 1\}$ and compare the model performance under these settings. As illustrated in Fig.~\ref{fig:single_multi_round} (a), the performance of the searched model is relatively stable when $\lambda$ ranges from 0.7 to 0.9. When $\lambda$ decreases from 0.7, the performance of the final pruned architecture gradually drops. The trend is expected since \algorithmname~only performs the evaluation of layer importance once on the over-pruned model. The importance evaluation would become inaccurate given a small $\lambda$, as more resources are assigned to the pool while the over-pruned backbone is too slim. Another phenomenon we observe is that as the pruning becomes more aggressive, the performance of the pruned model is more susceptible to the value of $\lambda$ (see the green line in Fig.~\ref{fig:single_multi_round} (a)). We conjecture that the performance drop is caused by the challenging pruning condition. Given very limited FLOPs resources, one has to allocate resources in a very careful manner so that the pruned model can exhibit satisfactory performance.

\noindent \textbf{One-round reallocation v.s. multi-round reallocation:} Recall that we adopt the one-round resource reallocation in our algorithm, \ie, allocating the resources by just performing a single pass of layer importance estimation. An alternative strategy is to evenly divide the resources into several parts and reallocate these resources successively in multiple rounds. Multi-round reallocation is more expensive as one needs to repeat Step-2 and Step-4 iteratively (see Sec.~\ref{sec:methodology}), and each iteration involves the fine-tuning of the backbone. As depicted in Fig.~\ref{fig:single_multi_round}, the performance of the multi-round resource reallocation is more stable than the one-round version when the $\lambda$ is set 0.6 and 0.5. For instance, when the target FLOPs is 2.2G and $\lambda$ is set as 0.5, the performance of the one-round reallocation decreases from 76.8\% to 75.3\% while the multi-round reallocation can still achieve 76.0\% in top-1 accuracy. The more stable performance may come from the fact that multi-round resource reallocation reevaluates the importance of each layer in each round, thus allowing a more appropriate resource reallocation.

\noindent \textbf{Robustness to the layer importance indicator:} The original \algorithmname~adopts BN statistics to reflect the importance of each layer. Here, we explore different importance indicators and check whether the proposed \algorithmname~is sensitive to the chosen indicator. We choose the widely-used filter norm~\cite{li2017pruning} and reconstruction errors~\cite{luo2017thinet} as the criteria. Layers with large filter norm or large reconstruction errors are considered as important. However, directly taking filter norm as the evaluating metric is biased since the filter norm of different layers have diverse scales in magnitude~\cite{chin2020towards}. To correct such bias, we follow~\cite{chin2020towards} and learn the layer-wise affine transformations for the filter norm. Under this circumstance, filter norm can better reflect the value of each layer. From Table~\ref{diff_indicator_table}, when pruning ResNet-50 under different FLOPs requirements, architectures searched by \algorithmname~using BN statistics, filter norm and reconstruction errors achieve similar performance. The results suggest the robustness of \algorithmname, irrespective of the methods used in evaluating layer importance.  

\begin{table}[t]
\caption{Performance of \algorithmname~with different importance criteria on ResNet-50.}
\vskip 0.1cm
\label{diff_indicator_table}
\centering
\small{
\begin{tabular}{c|c|c|c}
\hline
\multirow{2}*{Criterion} & \multicolumn{3}{c}{$\#$FLOPs} \\
\cline{2-4}
~ & 2.2 G & 1.8 G & 1.1 G \\
\hline
\hline
BN statistics (Ours) & 76.8\% & 75.5\% & 75.1\% \\
\hline
Filter norm~\cite{li2017pruning} & 76.7\% & 75.2\% & 75.2\% \\
\hline
Reconstruction errors~\cite{luo2017thinet} & 76.6\% & 75.4\% & 75.0\% \\
\hline
\end{tabular}
}
\end{table}

\begin{figure}[t]
 \centering
 \includegraphics[width=1.0\linewidth]{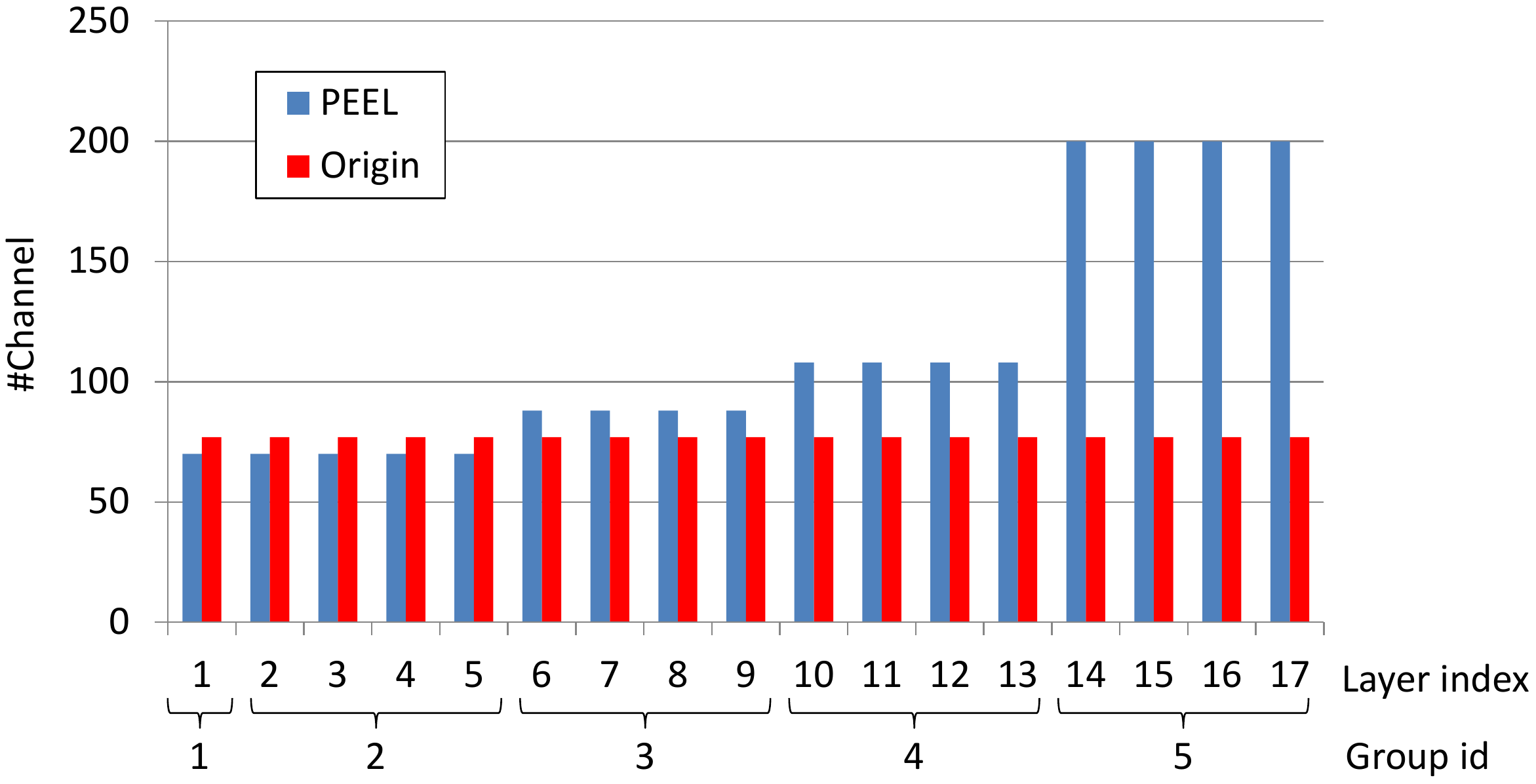}
 \vskip -0.1cm
 \caption{Visualization of the architecture found by \algorithmname~on a different backbone model that has the same number of channels for all layers. The base model is ResNet-18, and target FLOPs is 1.04G.}
 \centering
 \vskip -0.4cm
 \label{fig:diff_backbone}
\end{figure}

\noindent \textbf{Effect of knowledge distillation:} By comparing the two columns of the Top-1 Acc in Table~\ref{result_table}, we find that \algorithmname~consistently yields better performance than baseline pruning approaches with or without knowledge distillation (KD). For instance, when performing pruning on ResNet-18 and target FLOPs is 1.33G, \algorithmname~with KD is 0.6\% higher than DMCP with KD. One interesting phenomenon we observe is that KD brings more gains when the backbone model has fewer parameters. For example, KD can bring approximately 0.3\%, 0.5\% and 1.0\% to \algorithmname~on ResNet-50, ResNet-18 and MobileNetV2, respectively when the pruning is not aggressive. We conjecture that the increased gains are attributed to the deficiency of small models in grasping knowledge in labels by themselves and hence they more eagerly call for the guidance of the original model.

\noindent \textbf{Efficacy of \algorithmname~on a different backbone model:} Recall that we treat the uniformly trimmed model as the backbone and conduct resource reallocation on it. It is natural to wonder whether the resource reallocation would still work if a totally distinct architecture is used. Here, we use an architecture that has the same number of channels across all layers as the new starting point. 
All configurations are the same as the previous experiments. From Fig.~\ref{fig:diff_backbone}, we can observe that in this new backbone, groups 1, 2 and 3 are assigned with few parameters while the majority of the resources are distributed to groups 4 and 5. Overall, \algorithmname~puts more parameters in the deeper layers, and the resulting architecture is \textbf{8.7\%} higher in terms of top-1 accuracy in comparison to the original model (62.4\% v.s. 53.7\%). The result demonstrates the effectiveness of resource reallocation as well as the insensitivity of our algorithm to the backbone model.

\begin{figure}[t]
 \centering
 \includegraphics[width=1.0\linewidth]{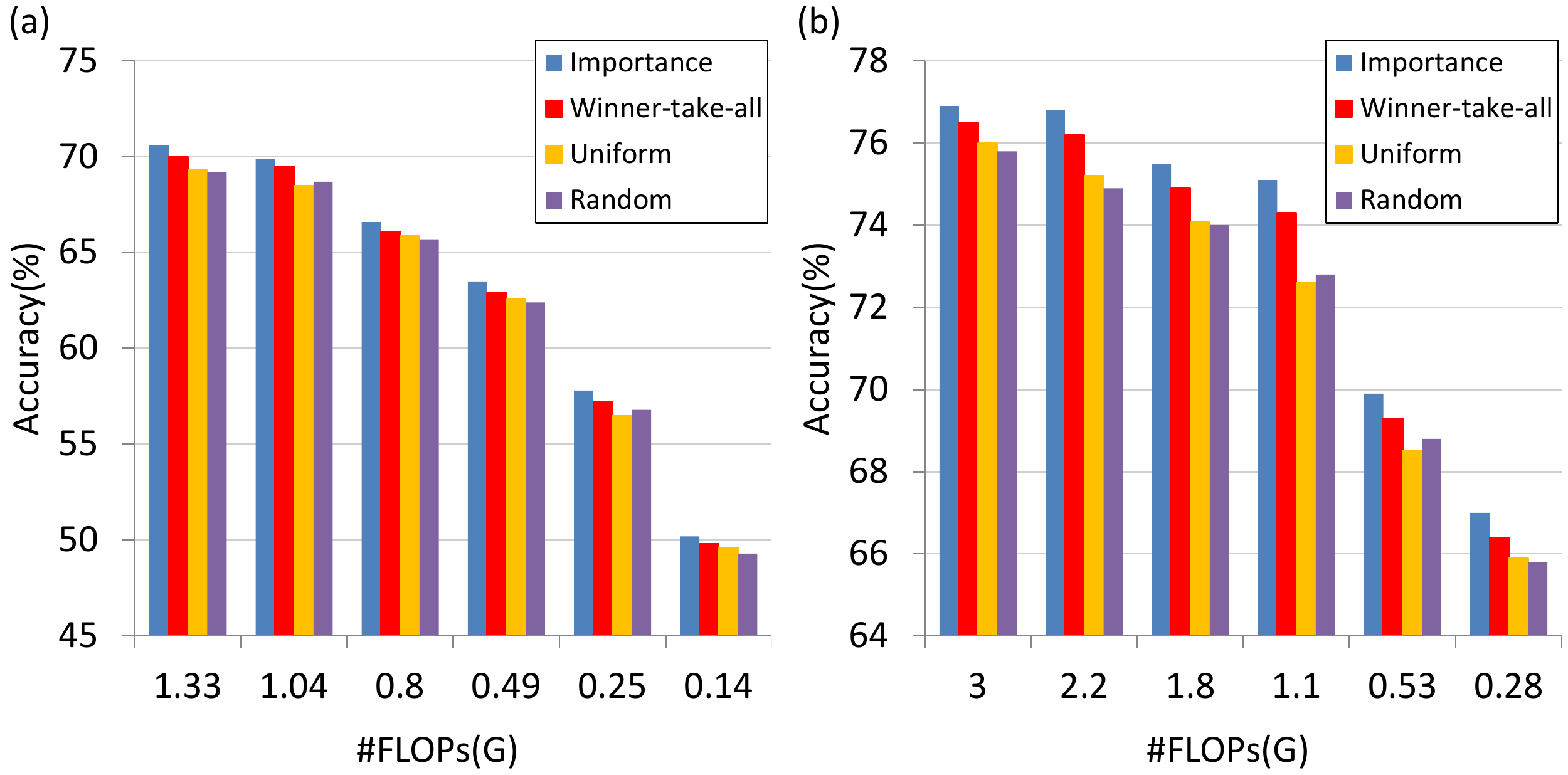}
 \vskip -0.1cm
 \caption{Policy on resource reallocation. Performance of \algorithmname~with importance-guided, winner-take-all, uniform and random policies on (a) ResNet-18 and (b) ResNet-50, respectively.}
 \centering
 \vskip -0.2cm
 \label{fig:diff_policy}
\end{figure}

\noindent \textbf{Policy on resource reallocation:} Here, we compare our `importance-guided' resource reallocation strategy with other parameter reallocation policies, \ie, winner-take-all, uniform and random reassignment. The winner-take-all policy determines the most important group according to the computed group significance and puts all resources in that group. 
The uniform policy adds the same number of channels to all layers while the random policy stochastically selects several layers and increases their channels. Here, we name the original reallocation strategy as importance-guided policy as it distributes resources based upon the estimated layer importance. From Table~\ref{fig:diff_policy}, our importance-guided reallocation policy evidently outperforms the other three policies on ResNet-18 and ResNet-50. For instance, when the target FLOPs is 1.8G on ResNet-50, the top-1 accuracy of the importance-guided policy is 75.5\% while the top-1 accuracy of the winner-take-all, uniform and random policy is 74.9\%, 74.1\% and 74.0\%, respectively. The superior performance demonstrates the effectiveness of the importance-guided policy.

\section{Conclusion}\label{conclusion}

We have presented an easy-to-implement yet effective channel pruning algorithm, \ie, \algorithmname, to acquire a desired compact model via reallocating resources from less informative layers to more crucial layers. The intuition of \algorithmname~is that layers do not contribute equally to the model performance and those having an indispensable influence on the performance should be assigned more resources to magnify their positive effect. We conduct extensive experiments to verify the efficacy of \algorithmname~on ImageNet dataset with modern CNN architectures, \ie, ResNet-18, ResNet-50, MobileNetV2, MobileNetV3-small and EfficientNet-B0. Experimental results suggest the effectiveness of \algorithmname~in uncovering compact yet accurate architectures consistently under various pruning levels, compared with state-of-the-art channel pruning methods. Besides, \algorithmname~yields stable searching results with small performance variance, and the searching cost is evidently smaller than contemporary channel pruning algorithms (\eg, DMCP~\cite{guo2020dmcp}). 

{\small
\bibliographystyle{ieee_fullname}
\bibliography{egbib}
}

\appendix
\appendixpage
\addappheadtotoc

\section{A detailed comparison with DMCP}

\begin{figure*}[!t]
 \centering
 \includegraphics[width=1.0\linewidth]{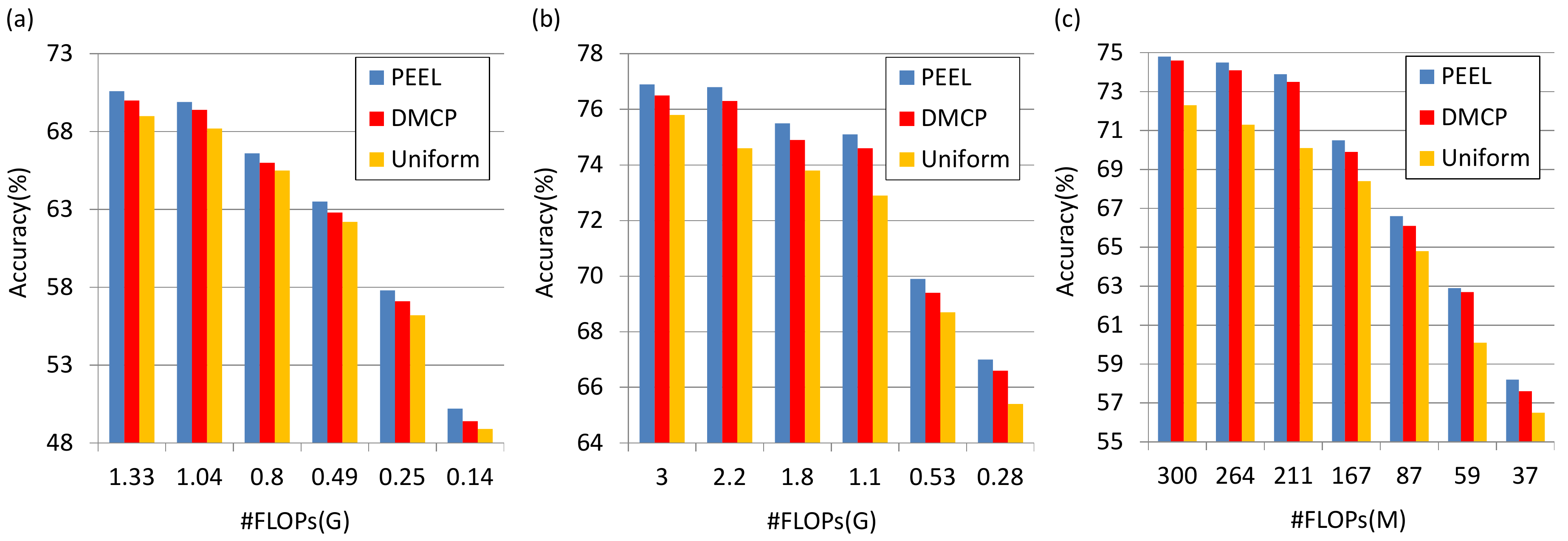}
 \vskip -0.1cm
 \caption{FLOPs-accuracy spectrum of \algorithmname, DMCP and uniform pruning on (a) ResNet-18, (b) ResNet-50 and (c) MobileNetV2.}
 \centering
 \vskip -0.2cm
 \label{fig:flops_acc_all}
\end{figure*}

We compare \algorithmname~against the recent and representative DMCP~\cite{guo2020dmcp}.
Figure~\ref{fig:flops_acc_all} shows the accuracy of architectures pruned by PEEL and DMCP under diverse FLOPs constraints on ResNet-18, ResNet-50 and MobileNetV2. Here, we take the uniform pruning as reference. It is observed that \algorithmname~can always find better architectures than DMCP in terms of top-1 accuracy. On all FLOPs levels, \algorithmname~outperforms DMCP by least 0.4\% in terms of top-1 accuracy. Thanks to the resource reallocation, the top-1 accuracy gap between \algorithmname~and uniform pruning lies between 1.0\% and 3.8\%. 

\begin{figure}[!t]
 \centering
 \includegraphics[width=0.6\linewidth]{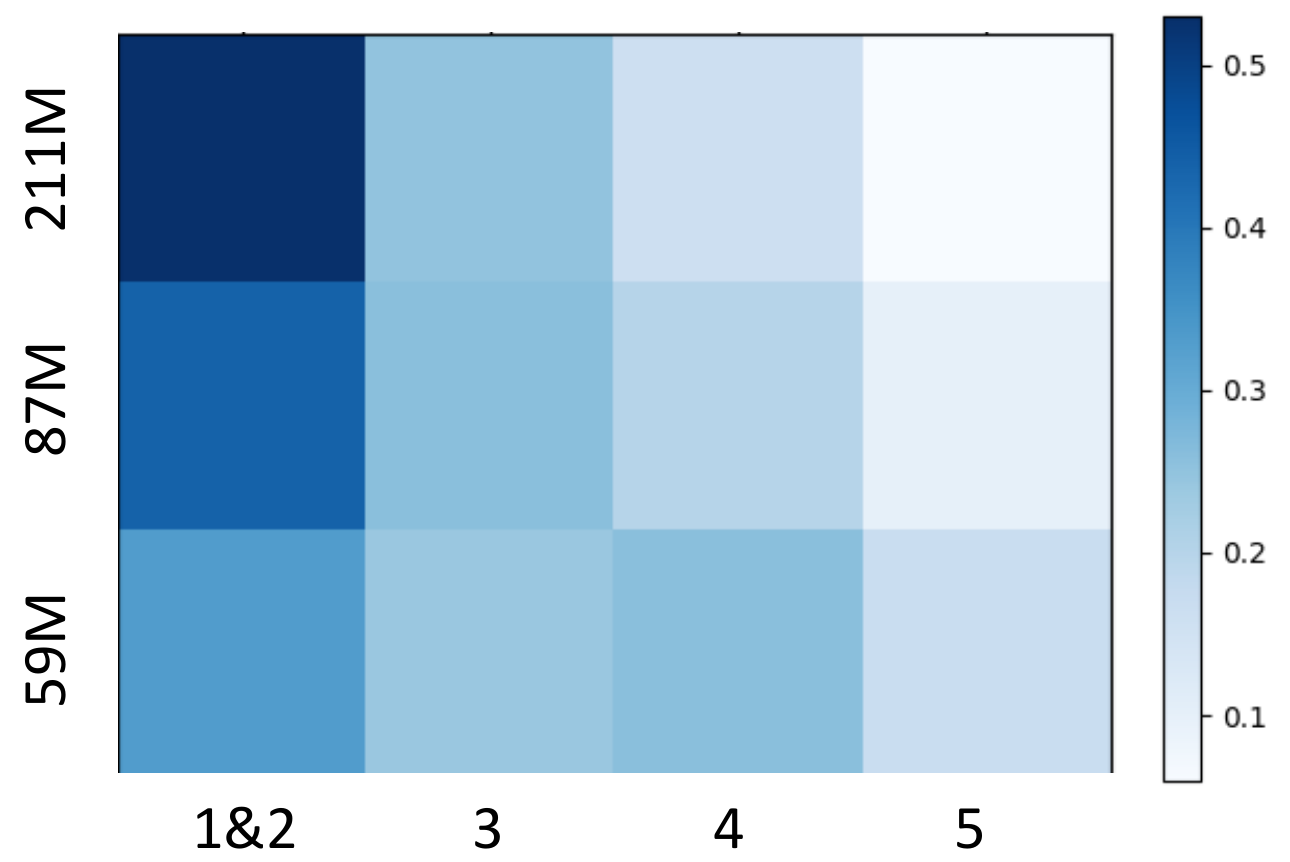}
 \vskip -0.1cm
 \caption{Percentage of FLOPs assigned by \algorithmname~to different groups of MobileNetV2 under different target FLOPs constraints. X-axis denotes group indices and y-axis represents the target FLOPs. A darker color indicates a larger quantity of assigned resources.}
 \centering
 \vskip -0.2cm
 \label{fig:flops_assign_mbv2}
\end{figure}

Figure~\ref{fig:flops_assign_mbv2} reveals the detailed FLOPs assignment on MobileNetV2~\cite{sandler2018mobilenetv2}. We observe a similar pattern in the FLOPs distribution process of MobileNetV2 to that of ResNet-18 and ResNet-50. Concretely, when the pruning is mild, the resource pool will tend to allocate more resources in the first three groups as more redundency is expected in deep layers. When the target FLOPs budget decreases, the reallocated resources will be more even for five groups as all groups are in need of more parameters to function effectively.


\begin{table}[t]
\caption{Performance variance of searched architectures of PEEL and DMCP on ResNet-50. Results are averaged over five runs.}
\vskip 0.1cm
\label{performance_var_table_supp}
\centering
\small{
\begin{tabular}{c|c|c|c}
\hline
\multirow{2}*{Method} & \multicolumn{3}{c}{$\#$FLOPs} \\
\cline{2-4}
~ & 2.2G & 1.8G & 1.1G \\
\hline
\hline
\textbf{\algorithmname~} & \textbf{76.7\% $\pm$ 0.2\%} & \textbf{75.3\% $\pm$ 0.3\%} & \textbf{75.0\% $\pm$ 0.2\%} \\
\hline
DMCP~\cite{guo2020dmcp} & 75.7\% $\pm$ 0.7\% & 74.2\% $\pm$ 0.8\% & 73.8\% $\pm$ 0.8\% \\
\hline
USNet~\cite{yu2019universally} & 75.4\% $\pm$ 0.6\% & 74.0\% $\pm$ 0.7\% & 73.4\% $\pm$ 0.6\% \\
\hline
\end{tabular}
}
\end{table}

\begin{table}[!t]
\caption{Comparison between the training time and memory usage of \algorithmname~and DMCP on ResNet-50 ($\#$FLOPs=1.1G). Here, the memory usage is measured on one GPU and the batch size is set as 64 for each GPU.}
\vskip 0.1cm
\label{train_cost_table_supp}
\centering
\small{
\begin{tabular}{c|c|c}
\hline
Algorithm & Train Time (H) & Memory usage (G) \\
\hline
\hline
\textbf{\algorithmname~} & \textbf{37} & \textbf{4.3} \\
\hline
DMCP~\cite{guo2020dmcp} & 98 & 9.5 \\
\hline
USNet~\cite{yu2019universally} & 66 & 8.7 \\
\hline
\end{tabular}
}
\end{table}

\begin{figure}[t]
 \centering
\subfloat[ResNet-18]{\includegraphics[width=1.0\linewidth]{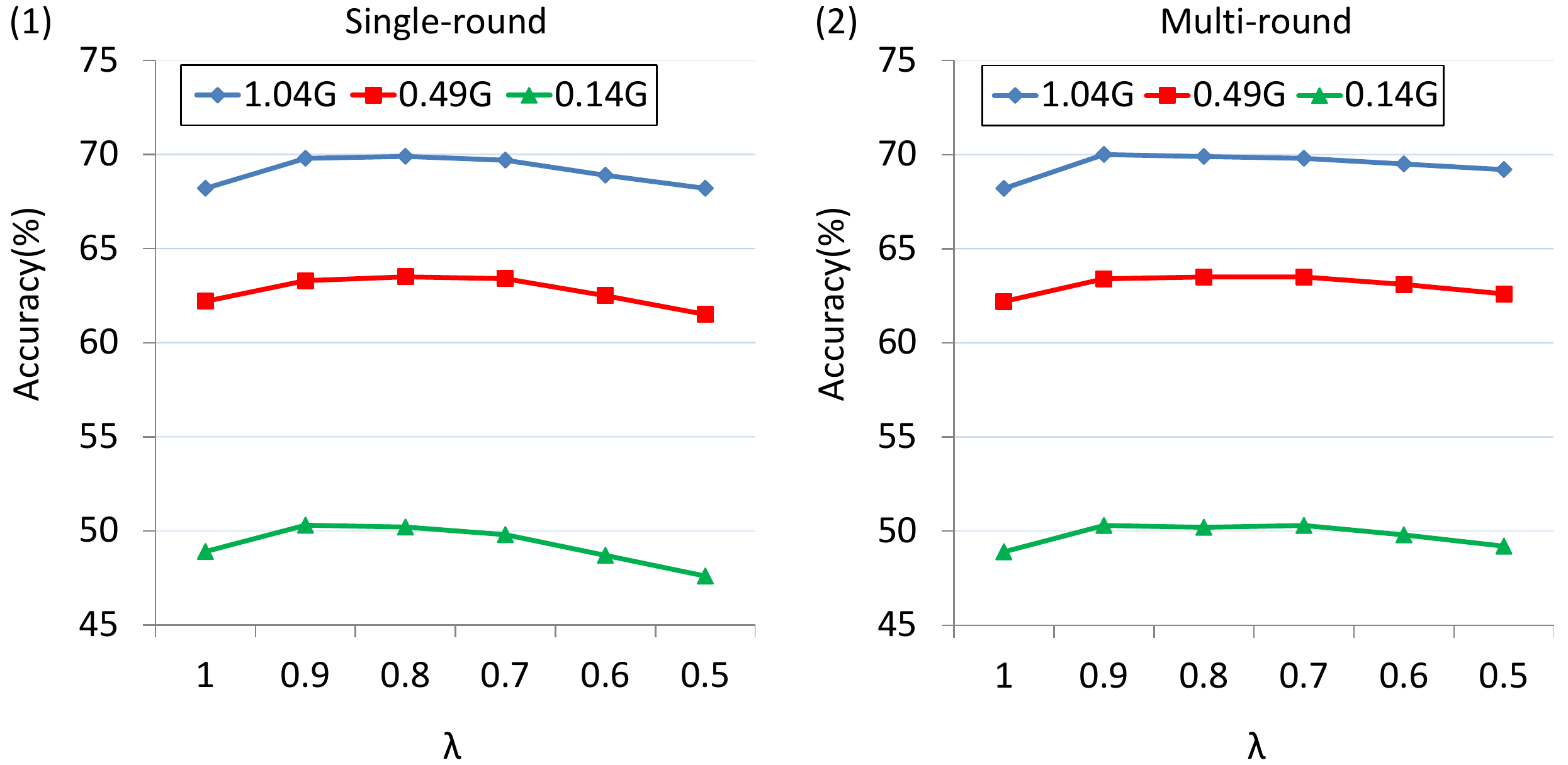}}\\
\subfloat[MobileNetV2]{\includegraphics[width=1.0\linewidth]{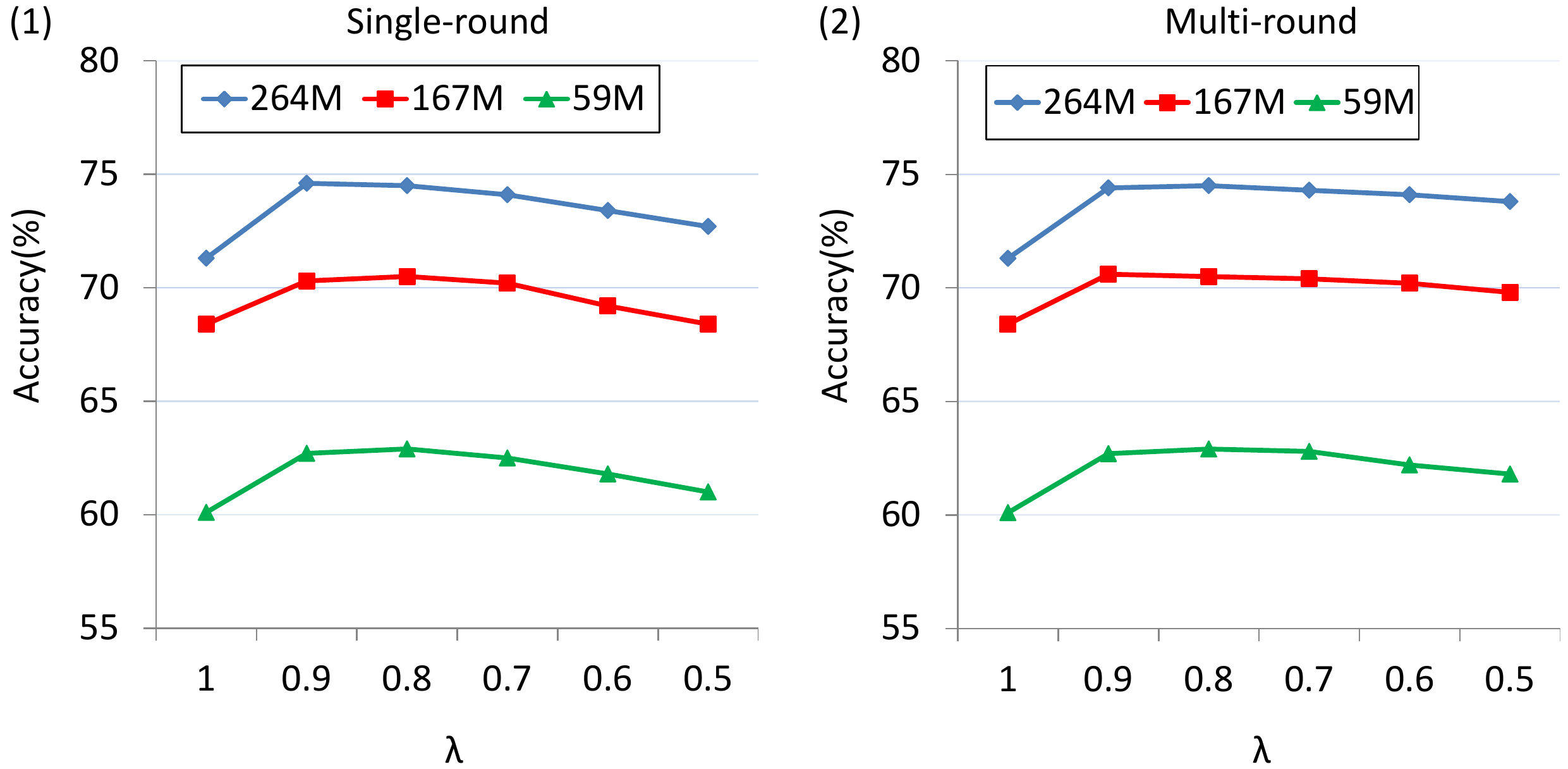}}
 \caption{One-round v.s. multi-round reallocation. We show the $\lambda$-accuracy spectrum of one-round (left) and multi-round (right) resource reallocation of \algorithmname~on (a) ResNet-18 and (b) MobileNetV2, respectively. Different color denotes different target FLOPs.}
 \centering
 \vskip -0.2cm
 \label{fig:single_multi_round_r18_mbv2}
\end{figure}

\section{$\lambda$-accuracy spectrum of one-round and multi-round resource reallocation}

We have shown the results of ResNet-50 with one-round and multi-round reallocation in Figure 6 of the main paper. And here, we provide detailed results on ResNet-18 and MobileNetV2. It is noteworthy that we observe a similar trend on ResNet-18 and MobileNetV2. Specifically, as shown in Fig.~\ref{fig:single_multi_round_r18_mbv2} (1), \algorithmname~with one-round reallocation can produce relatively stable results when $\lambda$ is not smaller than 0.7. When $\lambda$ continues to decrease, the performance of \algorithmname~suffers as the one-round evaluation of layer importance becomes less accurate. \algorithmname~with multi-round resource reallocation (Fig.~\ref{fig:single_multi_round_r18_mbv2} (2)) can avoid severe performance drop when $\lambda$ is set as 0.6 or 0.5 since it performs the importance assessment during each round, thus yielding more reliable statistics of layer importance.

\begin{figure*}[t]
\centering
\subfloat[ResNet-50]{\includegraphics[width=0.8\linewidth,height=8cm,keepaspectratio]{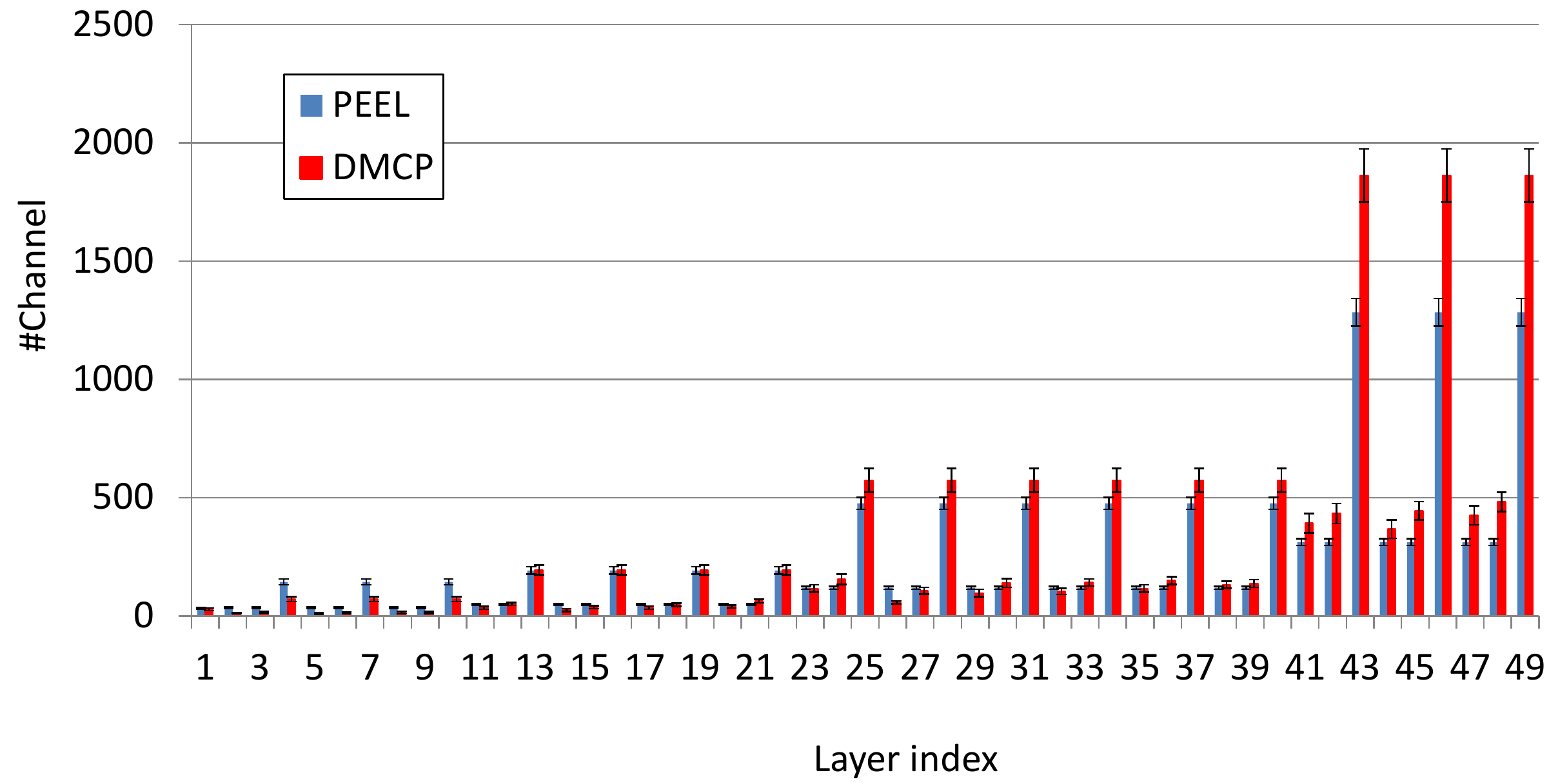}}\\
\subfloat[MobileNetV2]{\includegraphics[width=0.8\linewidth,height=8cm,keepaspectratio]{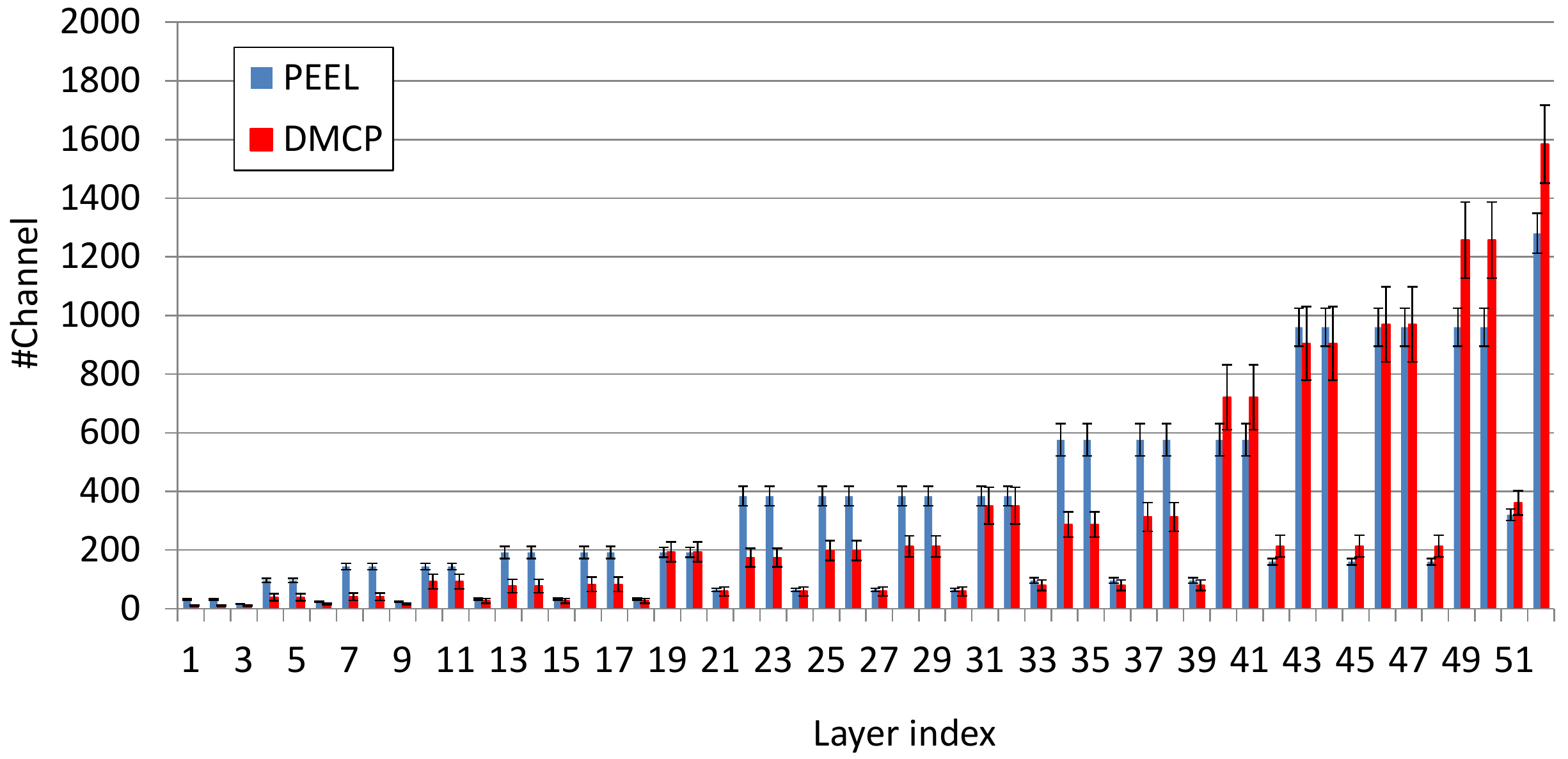}}
 \caption{Comparison between structures searched by \algorithmname~and DMCP on (a) ResNet-50 when $\#$FLOPs=1.1G and (b) MobileNetV2 when $\#$FLOPs=300M, respectively. The vertical bar and vertical line denote the mean and variance of the channel numbers in each layer, respectively. Results are averaged over five runs.}
 \label{fig:arch_compare_r50_mbv2}
\end{figure*}

\section{Visualization of architectures searched by \algorithmname~and DMCP}

We visualize the architectures obtained by \algorithmname~and DMCP~\cite{guo2020dmcp} on ResNet-50~\cite{he2016deep} and MobileNetV2~\cite{sandler2018mobilenetv2}. As depicted in Fig.~\ref{fig:arch_compare_r50_mbv2}, on both ResNet-50 and MobileNetV2, \algorithmname~puts more resources on shallow layers whilst DMCP has more reallocated parameters on the deep layers. As to the architectural stability, the variance of channel numbers of \algorithmname~is much smaller than that of DMCP. For instance, on ResNet-50, the channel variance of the last nine layers of \algorithmname~is less than half of DMCP (28.8 v.s. 64.3). These results strongly support the searching stability of \algorithmname~.

\section{Comparison between performance variance and searching cost of \algorithmname~, DMCP and USNet}

We summarize the performance variance and searching cost of different pruning algorithms in Table.~\ref{performance_var_table_supp} and Table.~\ref{train_cost_table_supp}, respectively. It is evident that \algorithmname~has less variance in performance than DMCP~\cite{guo2020dmcp} and USNet~\cite{yu2019universally}. Besides, \algorithmname~takes shorter training hours and consumes much less GPU memory to produce a desired slim model than the other two pruning techniques. It is not surprising since \algorithmname~trains a much compact model and reallocates resources on this model while both DMCP and USNet entails the training of the original cumbersome network. Besides, as opposed to DMCP and USNet, \algorithmname~is free from calculating the gradients of multiple sampled substructures, thus occupying fewer memory resources during the searching phase. These results explicitly showcases the superior efficiency of \algorithmname.

\end{document}